\documentclass{article}
\usepackage{arxiv}

\usepackage{palatino, epsfig,latexsym}
\usepackage{amsthm}
\usepackage{graphicx}
\usepackage{tikz}
\usepackage{pgfplots}
\usepackage[T1]{fontenc}
\usepackage{lmodern}
\usepackage{pdflscape}
\newcommand{\ignore}[1]{}
\usetikzlibrary{matrix}
\usepgfplotslibrary{groupplots}


\usepackage{amsmath}
\usepackage{mathtools}
\usepackage{wasysym}

\usepackage{tikz}
\usepackage{tkz-euclide}
\usepackage{xspace}
\usepackage{epsfig}
\usepackage{color}
\usepackage{url}
\usepackage{graphicx,dsfont}
\usepackage{tikz}
\usepackage{float}
\usepackage{longtable}
\usepackage{booktabs}
\usepackage{makecell}
\usepackage{multicol}
\usepackage{multirow}
\usepackage{colortbl}
\usepackage{csquotes}
\usepackage{algorithm}
\usepackage[noend]{algorithmic}
\usepackage{soul}

\usepackage{tikz}
\usepackage{pgfplots}
\pgfplotsset{compat=1.17}

\definecolor{ashgrey}{rgb}{0.7, 0.75, 0.71}
\definecolor{mygray}{rgb}{0.86, 0.86, 0.86}

\sethlcolor{mygray}

\DeclareMathOperator*{\argmax}{arg\,max}

\usepackage{xcolor}
\usepackage{amssymb}

\definecolor{todocolor}{rgb}{0.9,0.1,0.1}
\definecolor{changedcolor}{rgb}{0.42,0.27,0.57}
\definecolor{addedcolor}{rgb}{0.867,0.176,0.361}

\newcommand{\redacted}[1]{\emph{[anonymized for submission]}}
\renewcommand{\redacted}[1]{#1}

\begin{document}

% \ecjHeader{x}{x}{xxx-xxx}{201X}{Entropy-Based EDO for the TSP} {Nikfarjam et al.}
\title{On the Use of Survival Selection Methods for Evolutionary Diversity Optimisation}
\author{
 Adel Nikfarjam \\
Optimisation and Logistics\\School of Computer Science\\ Adelaide University\\
  \texttt{adel.nikfarjam@adelaide.edu.au} \\
  %% examples of more authors
   \And
 Jakob Bossek \\
Department of Computer Science\\Machine Learning and Optimisation\\Paderborn University\\
  \texttt{jboss@mail.uni-paderborn.de} \\
  \And
 Aneta Neumann \\
Optimisation and Logistics\\School of Computer Science\\Adelaide University\\
  \texttt{aneta.neumann@adelaide.edu.au} \\
    \And
 Frank Neumann \\
Optimisation and Logistics\\School of Computer Science\\Adelaide University\\
  \texttt{frank.neumann@adelaide.edu.au} \\}

% \author{
%         \name{\bf A. Nikfarjam} \hfill \addr{adel.nikfarjam@adelaide.edu.au}\\
%         \addr{School of Computer and Mathematical Sciences, The University of Adelaide,
%         Adelaide, Australia}
%   %      \addr{Department of Computer Science, The University of Adelaide,
%   %      Adelaide, Australia}
% \AND
%         \name{\bf J. Bossek} \hfill \addr{bossek@aim.rwth-aachen.de}\\
%         \addr{Department of Computer Science, RWTH Aachen University, 
%         Aachen, Germany}
% \AND'
%        \name{\bf A. Neumann} \hfill \addr{aneta.neumann@adelaide.edu.au}\\
%                \addr{School of Computer and Mathematical Sciences, The University of Adelaide,
%         Adelaide, Australia}
%  %       \addr{Department of Computer Science, The University of Adelaide,
% %        Adelaide, Australia}
% \AND
%        \name{\bf F. Neumann} \hfill \addr{frank.neumann@adelaide.edu.au}\\
%                \addr{School of Computer and Mathematical Sciences, The University of Adelaide,
%         Adelaide, Australia}
%  %       \addr{Department of Computer Science, The University of Adelaide, 
% %        Adelaide, Australia}
% }

\maketitle
\thispagestyle{plain}
\pagestyle{plain}
\begin{abstract}
Generating a diverse set of high quality solutions for an optimisation problem has been studied extensively in recent years by the evolutionary computation community. 
A paradigm that has received increasing attention is evolutionary diversity optimisation (EDO), where the goal is to maximise the diversity of a solution set subject to quality constraints. Since the contribution of each solution to the diversity of the population depends on other solutions and can change dramatically if several solutions in the population are modified simultaneously, most EDO approaches generate a single new solution per generation and discard the solution with the least contribution to diversity, ensuring a steady increase in population diversity over successive generations until convergence.
In this study, we aim to answer two questions: (1) Is generating multiple solutions in each generation beneficial for EDO? (2) How can this be achieved efficiently, given that conventional survival selection methods do not work well in EDO due to the dependency of a solution's contribution to diversity on other solutions? 
\end{abstract}

% \keywords{
%     Evolutionary diversity optimisation,
%     travelling salesperson problem,
%     high-order entropy,
%     diversity subset selection
% }

% \frank{page limit: 6 pages (plus 2 at additional charge if required)}
% \frank{cut paper to 6 pages, decrease figures and tables}

% \frank{Would be good to check
% Yu-Ran Gu, Chao Bian, Miqing Li, Chao Qian:
% Subset Selection for Evolutionary Multiobjective Optimization. IEEE Trans. Evol. Comput. 28(2): 403-417 (2024)
% and see how it relates to our paper.
% We could also change our title to "Subset selection for EDO"} 

\section{Introduction}

%In most optimisation problems, the goal is to find (near-)optimal solutions with respect to one or more objectives.
The standard goal in single-objective optimisation is to find one single solution that is optimal or of high quality.
Over the last decade, there has been a significant rise in the number of studies seeking a diverse set of high quality solutions for a given single-objective optimisation problem. Having a diverse set of high quality solutions can be valuable for a wide range of reasons, such as providing information about the solution space, offering decision-makers a variety of diverse alternatives, and improving robustness against imperfect modelling and dynamic changes of the problem.

Traditionally, evolutionary computation (EC) research considers diversity as a means to escape local optima and prevent premature convergence, or to explore niches in the fitness landscape, which is often referred to as multimodal optimisation. Niching has been studied extensively in the literature; interested readers are referred to \cite{LiEDE17} for a review of niching methods. In recent years, quality diversity~(QD) and evolutionary diversity optimisation~(EDO), have emerged in the EC literature.

QD explores a predefined behavioural space and seeks the best-performing solutions in different regions of this space. The framework was first proposed in \cite{CullyM13}, while an algorithm to depict the distribution of high-performing solutions in a behavioural space was introduced simultaneously in \cite{CluneML13}. Later, the framework was formalised and named in \cite{PughSSS15, PughSS16}. QD has been studied in various domains such as combinatorial optimisation problems \cite{NikfarjamDN22, Bossek022}, improvement of health outcome through time-use plans \cite{NikfarjamSND024}, conditional search-space problems \cite{baraton2025bayesian}; however, most QD studies focus on robotics and video games \cite{TjanakaFTN22, EarleSFNT22}.

On the other hand, EDO maximises the structural diversity of a fixed-size set of solutions, subject to solution quality. This means that the objective function of the given optimisation problem is treated as a constraint, while the diversity measure plays the role of the fitness function. Although EDO was first introduced in a continuous domain \cite{ulrich2011maximizing}, it has been studied extensively in the context of combinatorial optimisation in recent years. Several studies have incorporated EDO to generate diverse sets of travelling salesperson problem (TSP) instances and images \cite{alexander2017evolution, doi:10.1162/evcoa00274, neumann2018discrepancy, neumann2019evolutionary, bossek2019evolving}. There have also been several papers employing EDO to generate diverse solutions for different combinatorial problems. EDO has been utilised to generate diverse set of solutions for many optimisation problems: travelling salesperson problem~(TSP)~\cite{viet2020evolving, NikfarjamBN021a, NikfarjamB0N21b},
% In \cite{viet2020evolving}, the authors employed two distance-based diversity measures to calculate pairwise edge overlaps and used them as the fitness function. The focused was on an entropy-based measure in \cite{NikfarjamBN021a} that determines pairwise segment overlaps within the population, where the segment size can include one or more edges. The previous two studies assumed that an optimal tour (solution) is known a priori. In contrast, introduced a framework was introduced in \cite{NikfarjamB0N21b} that simultaneously searches for optimal solutions while diversifying the population in the context of EDO, and proposed an EAX-based crossover operator for this purpose.   
Quadratic Assignment Problem \cite{DoGNN22}, the minimum spanning tree problem \cite{Bossek021tree}, monotone sub-modular functions \cite{NeumannB021}, the travelling thief problem \cite{NikTTPEDO},
% They scrutinised the inter-dependency of TTP sub-problems in terms of diversity and highlighted its affects on the diversity of solutions. An EDO approach was recently adopted to effectively provide diverse sets of high-quality
% solutions for 
the detection and concealment of communication networks in large settings~\cite{neumannGECCO23}, and  the boolean satisfiability problem \cite{NIKSAT}. In most of these studies, a vanilla $(\mu+1)$~EA has been utilised to maximise diversity, where at most one solution was replaced in each generation. Removing the individual with smallest contribution to diversity is used in \cite{neumannGECCO23} who consider $(\mu+\lambda)$~EA with $\lambda$ up to $10$. It is worth mentioning that subset selection has gained increasing attention within the evolutionary multi-objective optimisation (EMO) community in recent years~\cite{GuBLQ24}. 

\subsection{Our Contribution}

Most studies in the EDO literature work with a $(\mu+1)$~EA, where a single solution is generated in each iteration. A solution with the least contribution to the population diversity is then discarded, and the EA proceeds to the next generation. 
We investigate EDO algorithms with an offspring population size greater than $1$ and explore its potential benefit.
%The underlying reason is that the contribution of a solution to diversity depends on the other solutions in the population. Therefore, if a $(\mu+\lambda)$~EA is used, $\lambda$ solutions are generated in each iteration and $\lambda$ solutions need to be discarded to move to the next generation. However, due to the dependency of solutions’ diversity contributions on each other, conventional survival selection methods do not work, and subset selection for the next generation is non-trivial.
Accordingly, we aim to answer the following two questions: (1) How does the use of a $(\mu+\lambda)$~EA affect the diversity of solutions? (2) What is an efficient way to select the next generation in EDO? To answer these questions, we introduce two survival selection methods: (1) a tournament-based subset selection method, where subsets of solutions from the offspring and the parent population are iteratively selected and compete with each other, and the solution with the least contribution is discarded until $\mu$ solutions remain; the key point here is that after each deletion, the contribution of the remaining solutions must be updated, and (2) an EA-based subset selection method, where an inner EA is introduced to determine the next generation; we compare these methods with the greedy approach in \cite{neumannGECCO23} where the solutions
with the least contribution are removed one by one until $\mu$ solutions remain. 
%Moreover, previous EDO methods tend to lose solution quality rapidly until the solutions approach the quality constraint boundary; at that point, there is little room left to modify the solutions in order to further increase diversity and the population is stuck in local diversity optima. 
We incorporate a mechanism into the EA-based selection method that takes into account the length of solutions in addition to their contribution to diversity, helping to avoid local diversity optima.
We conduct a comprehensive empirical study to compare these methods with the conventional $(\mu+1)$~EA used in the literature. Experiments show that EA-based selection outperforms the other methods in most instances.

The remainder of this paper is structured as follows: Section \ref{Sec:define} defines EDO and the TSP problem as the benchmark case study. Section \ref{Sec:mu_p_lam} presents the $(\mu+\lambda)$~EA and introduces three survival selection approaches for the algorithm. \ref{Sec:EXp_lam} compares $(\mu+\lambda)$~EA using the selection methods and the conventional $(\mu+1)$~EA thorough a comprehensive experimental investigation. We conclude with some remarks and future research perspectives. 

\section{Preliminaries}
\label{Sec:define}
We consider the TSP as the benchmark problem in this study. The TSP is defined on a directed complete graph $G = (V, E)$ where $V$ is a set of $n =|V|$ nodes and $E$ is the set of pairwise edges between the nodes. A non-negative distance (weight) $w(e)$ is associated with each edge $e \in E$. The objective of the TSP is to find a roundtrip tour that visits all the nodes exactly once and returns to the first city (formally, a permutation $\pi = (\pi(1), \ldots, \pi(n))$ of the nodes is sought), with minimum distance travelled:
$$c(\pi) = w(\pi(n),\pi(1)) + \sum_{i=1}^{n-1} w(\pi(i),\pi(i+1)) \to \min!$$

In this study we consider the symmetric TSP where $w(u,v) = w(v,u)$ holds for all $u, v \in V$. As mentioned, EDO treats the actual objective function of the problem as constraints and employs a diversity measure as the fitness function:
%This is formulated as follows in the context of EDO:
\begin{align*}
    \max D(P)
    \text{ subject to }
    c(\pi) \leq (1+\alpha) \cdot \text{OPT}
    \,\,\forall \pi \in P.
\end{align*}
Here, $P$ is a $\mu$-size population of TSP tours, $D(P)$ is a measure to quantify the diversity of the population, $\text{OPT}$ is the objective value of an optimal solution for the TSP problem, $\alpha$ denotes a threshold for acceptable objective values for the TSP solutions. 
%The objective is to compute a set of solutions where all solutions have a length less than $(1+\alpha) \cdot \text{OPT}$ and  at the same time maximising the diversity of solutions with the respect to a diversity measure. 

\subsection{Diversity Measure}
EDO requires a measure to quantify the diversity of the TSP solutions. Here, we adopt the \emph{entropy} for two reasons: 1) it yields stronger results in TSP benchmark instances compared to distance-based diversity measures~\cite{nagata2020high}. 2) it has been incorporated in several EDO papers \cite{NikfarjamBN021a, NikfarjamB0N21b, NIKSAT}.
Let $f(e_{ij})$ be the number of tours in $P$ that contain the directed edge $e_{ij}$.
The entropy of a set of tours $P$ is calculated as
%\jakob{It is somehow confusing that the parameter $P$ does not apear on the right-hand side of the equation.}\adel{I changed a bit, not sure if it's better now!}

\begin{align*}
H(P) = -\sum_{e_{ij} \in P} \left(\frac{f(e_{ij})}{2n\mu}\right) \cdot\ln\left(\frac{f(e_{ij})}{2n\mu}\right)
\end{align*}

\ignore{
where
\begin{equation*}
h_P(e) = 
\begin{cases}
     0 & \mbox{if }f_P(e) = 0 %( e \notin P), 
     \\
     -\left(\frac{f_P(e)}{2n\mu}\right) \cdot\ln\left(\frac{f_P(e)}{2n\mu}\right)&
     \mbox{if }f_P(e) > 0 %(e \in P)
\end{cases}
\end{equation*}
is the entropy contribution of edge $e \in E$.
}
%In this equation $f(e)$ is the number of solutions in $P$ that include edge $e$. 
Note that each solution $\pi \in P$ includes $2n$ edges since we consider the symmetric TSP; given that, there are $\mu$ tours in $P$ and thus $P$ has $2n\mu$ edges in total.
Obviously, a population with the least diversity is when we have $\mu$ copies of a single tour in $P$, and the entropy value in this case is $H_{\min} = \ln(2n)$. 
% It has been shown in \cite{NikfarjamBN021a} that all edges should appear roughly the same number of times in the population to achieve maximum entropy. Therefore, based on the pigeonhole principle, if $\frac{2n\mu}{|E|}$ is an integer, all edges $e \in E$ must appear exactly $\frac{2\mu}{n-1}$ times; otherwise, some edges must appear $\left\lfloor \frac{2\mu}{n-1} \right\rfloor$ times and the remaining edges $\left\lfloor \frac{2\mu}{n-1} \right\rfloor + 1$ times. Let us denote $f_0 = \left\lfloor \frac{2\mu}{n-1} \right\rfloor$. 
The maximum entropy can then be calculated as follows:
\begin{align*}\nonumber
    H_{\max} = 
    & -((2n\mu)-(f_0\cdot |E|)) \cdot \left(\frac{f_0+1}{2n\mu}\right)\cdot \ln{\left(\frac{f_0+1}{2n\mu}\right)} \\ 
    & - ((f_0+1)\cdot |E|) -(2n\mu))\cdot \left(\frac{f_0}{2n\mu}\right)\cdot \ln{\left(\frac{f_0}{2n\mu}\right)} 
\end{align*}
Interested readers are referred to \cite{NikfarjamBN021a} for the proof and further details on $H_{\max}$.
\section{EDO with a $(\mu+\lambda)$~EA}
\label{Sec:mu_p_lam}
Most studies in the EDO literature use a $(\mu+1)$~EA to maximise the diversity of tours, where the offspring can replace either the parent or one of the individuals in the population $P$~(see, e.g., \cite{Bossek021tree} or \cite{NeumannB021}) since an increase in offspring pool raises the complexity of the problem due to dependency of diversity contribution of the solutions. 
% Unlike classical optimisation problems, where the fitness of a solution is independent of other solutions, diversity optimisation problems require the calculation of diversity values for a set of solutions, making the fitness of solutions dependent on other individuals in the population. 
% This increases the complexity of the problem by adding a subset selection sub-problem: how do we select a subset of $\mu$ individuals from a set of $\mu+\lambda$ individuals efficiently? Nevertheless, increasing the size $\lambda$ of the children's pool by using a $(\mu+\lambda)$~EA can potentially boost the performance of the EA if a suitable method for subset selection is incorporated into the algorithm.
This section aims to answer two questions: 1) Is it beneficial to increase the pool of offspring? 2) Given the complexity of the diversity problem, what is the best survival selection approach to determine the next generation? To answer these questions, we propose three subset selection methods to determine the next generation in each iteration of the EDO $(\mu+\lambda)$~EA.

\begin{algorithm}[t]
\begin{algorithmic}[1]
\REQUIRE{Population size $\mu$, size of offspring pool $\lambda \geq 1$}
\STATE Initialise the population $P$ with $\mu$ TSP tours such that\newline{}$c(\pi) \leq (1+ \alpha)\cdot \text{OPT}$ for all $\pi \in P$.\\
\STATE Set $Q = \emptyset$.
\WHILE{termination condition not met}
\WHILE{$|Q| < \lambda$}
\STATE Choose $\pi \in P$ uniformly at random and produce an offspring $\pi'$ of $\pi$ by mutation.\\ 
\STATE If $c(\pi') \leq (1+ \alpha)\cdot \text{OPT}$, add $\pi'$ to $Q$.\\
\ENDWHILE
\STATE Set $P = \text{\texttt{SELECT}}(P \cup Q)$. \hfill \COMMENT{Subset selection}\\
\ENDWHILE
%\STATE Repeat steps 2 to 6 until a termination criterion is reached.
\end{algorithmic}
\caption{Diversity maximising $(\mu+\lambda)$~EA}
\label{alg:ea_mu_plus_lambda}
\end{algorithm}

Algorithm~\ref{alg:ea_mu_plus_lambda} shows the pseudo-code of the EDO $(\mu+\lambda)$~EA with a parameterisable subset selection procedure (see call to function \texttt{SELECT} in line~7 of Algorithm~\ref{alg:ea_mu_plus_lambda}). In this study, we employed 2-opt as the mutation operator. 
The main challenge is, given a population $P$ with $|P| = \mu$, a set of offspring individuals $Q$ with $|Q| = \lambda \geq 1$, to select a subset $S^{*}$ of the union set $S = P \cup Q$ such that
\begin{align*}
S^{*} = \argmax_{S' \subset S, |S'| = \mu} H(S \setminus S').
\end{align*}
There are $\binom{\mu + \lambda}{\mu}$ possible subsets. As a consequence, iterating over all these subsets is computationally infeasible as the number of subsets grows exponentially if $\lambda$ is in the regime of $\varepsilon n$.
We therefore tackle this subset-selection problem by means of three different heuristics.

\begin{algorithm}[t]
\begin{algorithmic}[1]
\REQUIRE{Multi-set $S$ with $|S| = \mu + \lambda$}
\WHILE{$|S| > \mu$}
    \STATE Remove one individual $\pi$ from $S$, where $\pi = \argmax_{q \in S} H(S \setminus \{q\})$.
\ENDWHILE
\end{algorithmic}
\caption{Greedy subset selection.}
\label{alg:greedy_selection}
\end{algorithm}
\subsection{Greedy selection} The greedy selection strategy (see Algorithm~\ref{alg:greedy_selection}) is a straight-forward generalisation of the greedy survival selection that has already been used in \cite{neumannGECCO23}. In the $i$th iteration the algorithm removes an individual from the set $S$ whose deletion leads to smallest loss in diversity. Note that this selection method is computationally costly. It requires, assuming that all $\lambda$ offspring individuals adhere to the quality constraint, $(\mu+\lambda-i)$ $H$-evaluations in the $i$th iteration for $i = 0, \ldots, \lambda-1$ and thus the overall number of $H$-evaluations is
\begin{align*}
    \sum_{i=0}^{\lambda-1} (\mu + \lambda - i)
    %& = \sum_{i=1}^{\lambda} (\mu + i)
    %= \lambda \mu + \frac{\lambda(\lambda+1)}{2} \\
    = \lambda \left(\frac{2\mu + \lambda + 1}{2}\right)
    = \Theta(\lambda \mu + \lambda^2).
\end{align*}

\begin{algorithm}[t]
\begin{algorithmic}[1]
\REQUIRE{Multi-set $S$, $|S| = \mu + \lambda$, tournament size $r \geq 2$}
\WHILE{$|S| > \mu$}
    \STATE Select a subset $R \subseteq S$ of $r$ individuals uniformly at random with replacement.
    \STATE Remove one individual $\pi$ from $S$, where $\pi = \argmax_{q \in R} H(S \setminus \{q\})$.
\ENDWHILE
\end{algorithmic}
\caption{Tournament subset selection.}
\label{alg:ktournament_selection}
\end{algorithm}

Note that the $(\mu+1)$~EA adopted in most of EDO studies in the literature is a special case of the $(\mu+\lambda)$~EA in Algorithm~\ref{alg:ea_mu_plus_lambda} with the greedy selection method.

\subsection{Tournament selection} 
We adopt a tournament selection strategy iteratively (see Algorithm~\ref{alg:ktournament_selection}). In $\lambda$ iterations the algorithm samples a subset $R$ with $|R| = r$, $r \geq 2$ being the tournament size, individuals from the set uniformly at random with replacement. Eventually, the individual is dropped whose deletion yields the maximal $H$-value.
The number of $H$-evaluations is $\lambda \cdot r$ since the $H$-value is calculated $r$ times per iteration and it takes $\lambda$ iterations until the set is reduced to $\mu$ individuals; this is $\Theta(\lambda)$ for constant~$r$.

\begin{algorithm}[t]
\begin{algorithmic}[1]
\REQUIRE{Multi-set $S$ with $|S| = \mu + \lambda$}
\STATE Initialise $x \in \{0,1\}^{|S|}$ where $|x|_1 = \mu$.
% \STATE Let $H(x) = D(\{p_i \in P \mid x_i = 1\})$ be the diversity of the subset selected by $x$.
\FOR{$L$ iterations}
    % \STATE $L \gets 1 + \text{Poisson}(1)$\jakob{Suggestion to make the number of flipped bits random!} \adel{yes, good idea}\COMM{High(est) probability to flip exactly two bits.}
    \STATE Generate $y$ from $x$ by flipping one-bits independently with probability of $w/\mu$, where $w$ is selected based on power of law distribution. Then flip the same number of bits from zero to one randomly to maintain $|y|_1 = |x|_1 = \mu$.
    \IF[Refer to Eq.~\eqref{eq:div}]{$F(y) \geq F(x)$} 
    % \COMM{Refer to Equation \ref{eq:div}}
        \STATE Replace $x$ with $y$.
    \ENDIF
\ENDFOR
\end{algorithmic}
\caption{$(\mu+\lambda)$~EA-based subset selection.}
\label{alg:evolutionary_selection}
\end{algorithm}
\subsection{EA-based selection} The subset-selection process itself is optimised with a $(1+1)$~EA. To this end we maintain a bitstring $x \in \{0,1\}^{\mu + \lambda}$ and optimise the diversity of the set $S' = \{\pi_i \in S \mid x_i = 1\}$ under the equality constraint $|x|_1 = \mu$~(see Algorithm~\ref{alg:evolutionary_selection}). The algorithm is initialised such that exactly $\mu$ individuals are selected (i.e., $|x|_1 = \mu$). In $L$ iterations the algorithm generates an offspring $y$ by flipping zero-bits to ones and the same amount of one-bits to zero in the parent~$x$. The number of flipped bits is $2 \cdot l$ where $l$ is sampled from a Binomial distribution with parameters $\mu+\lambda$ and $\frac{w}{n} \in [0,1]$ where $w$ is selected based on a power-law distribution as it is done in the heavy-tailed mutation~\cite{DoerrLMN17}. This is done to allow the algorithm to escape local optima by increasing the probability to flip multiple ones and zeros. 
%\subsubsection{Length preservation mechanism for EA-based selection}
Similar to all EAs, the EA-based selection method requires a fitness function. The most obvious choice is to use the diversity of $y$ as the fitness function. However, focusing solely on $H(y)$ results in a rapid increase in solution length and increases the likelihood of getting stuck in local diversity optima. Therefore, we introduce a mechanism based on tour length to avoid stagnation in local optima. 
We first calculate $H_0$ and $l_0$ which are the diversity of the population $P$ and the summation of length of individuals in $P$, respectively. The fitness of offspring in Algorithm \ref{alg:evolutionary_selection} can be calculated from:
\begin{align}
F(x) =
\begin{cases}\label{eq:div}
    \frac{\Delta H }{\Delta l} & \mbox{if }  \Delta H > 0 \text{ and } \Delta l \geq 0 \\
	M & \mbox{if }  \Delta H \geq 0 \text{ and } \Delta l < 0 \\
    -M & \mbox{if }  \Delta H < 0
\end{cases}
\end{align}
where $\Delta H$ and $\Delta l$ are $H(y)- H_0$ and $l(y) - l_0$, respectively. We use $F$ as the fitness function, a higher $F$, a fitter solution. The first case is when the diversity of the population increased ($\Delta H > 0$) and the total length of the tours also increased or remained unchanged ($\Delta l \geq 0$) which means that the quality of the tours does not improve; then $F(y)$ is $\frac{\Delta H}{\Delta l}$. If the diversity of solutions increases or remains unchanged ($\Delta H \geq 0$) and the quality of the solutions improves ($\Delta l < 0$), then $F(y)$ is a large number $M$, and we break the search and choose the set $y$ for the next generation. Finally, if diversity decreases, $F(y)$ is equal to a large negative number $-M$ to ensure that we do not select a set that decreases diversity. Note that $L$ is set to $2\mu\lambda$ in this study based on preliminary investigations. Note that if $\Delta H \geq 0$ and $\Delta l < 0$, we stop the search for the iteration and go to the next generation. 

\section{Experimental Investigation}
\label{Sec:EXp_lam}
% Analogous to the experiments in Section~\ref{Sec:exp} 
We conduct two series of empirical investigations to study the impact of an offspring pool $\lambda \geq 2$. We first look at the algorithms' performance in the setting of unconstrained diversity optimisation. Then, we analyse constrained diversity optimisation. For the sake of fair comparison, we consider the number of diversity evaluations ($H$-evaluations) as the termination criterion. 

\subsection{Unconstrained Diversity Optimisation}
\begin{table*}
%\footnotesize
\centering
\caption{Comparison of the proposed subset selection schemes and $(\mu+1)$~EA in the unconstrained optimisation setting. The column \emph{rate} shows the number of times obtaining the optimal diversity out of 30 independent runs.  Stat shows the results of a Kruskal-Wallis test at significance level of $95\%$ with Bonferroni correction. $X^+$ means the median of the measure is better than the one for variant $X$, $X^-$ means it is worse and $X^*$ indicates no significant difference.}

\renewcommand{\tabcolsep}{6.5pt}
\renewcommand{\arraystretch}{0.9}
\begin{tabular}
{l|ccc|ccc|ccc|ccc}
\toprule
         &\multicolumn{3}{c|}{ $(\mu+\lambda)$~\textbf{EA}}&\multicolumn{3}{c|}{\textbf{Tournament}}&\multicolumn{3}{c|}{\textbf{Greedy}}&\multicolumn{3}{c}{$(\mu+1)$~\textbf{EA}}\\
\cmidrule(l{2pt}r{2pt}){2-4}
\cmidrule(l{2pt}r{2pt}){5-7}
\cmidrule(l{2pt}r{2pt}){8-10}
\cmidrule(l{2pt}r{2pt}){11-13}

             $\mu$& $H$ & Stat (1) & rate & $H$ & Stat (2) & rate & $H$ & Stat (3) & rate & $H$ & Stat (4) & rate \\

\midrule
20&7.6&$2^+3^+4^*$&\hl{\textbf{30}}&6.83&$1^-3^-4^-$&0&7.49&$1^-2^+4^-$&0&7.6&$1^*2^+3^+$&29\\
21&7.65&$2^+3^+4^*$&\hl{\textbf{30}}&6.86&$1^-3^-4^-$&0&7.53&$1^-2^+4^-$&0&7.65&$1^*2^+3^+$&18\\
22&7.7&$2^+3^+4^*$&\hl{\textbf{30}}&6.89&$1^-3^-4^-$&0&7.57&$1^-2^+4^-$&0&7.69&$1^*2^+3^+$&11\\
23&7.74&$2^+3^+4^+$&\hl{\textbf{30}}&6.92&$1^-3^-4^-$&0&7.6&$1^-2^+4^-$&0&7.73&$1^-2^+3^+$&0\\
24&7.78&$2^+3^+4^+$&\hl{\textbf{10}}&6.93&$1^-3^-4^-$&0&7.63&$1^-2^+4^-$&0&7.77&$1^-2^+3^+$&0\\
25&7.8&$2^+3^+4^+$&\hl{\textbf{16}}&6.96&$1^-3^-4^-$&0&7.65&$1^-2^+4^-$&0&7.78&$1^-2^+3^+$&0\\
26&7.78&$2^+3^+4^+$&\hl{\textbf{30}}&6.98&$1^-3^-4^-$&0&7.67&$1^-2^+4^-$&0&7.78&$1^-2^+3^+$&1\\
27&7.77&$2^+3^+4^*$&\hl{\textbf{30}}&7&$1^-3^-4^-$&0&7.69&$1^-2^+4^-$&0&7.77&$1^*2^+3^+$&24\\
28&7.76&$2^+3^+4^*$&\hl{\textbf{30}}&7.03&$1^-3^-4^-$&0&7.7&$1^-2^+4^-$&0&7.76&$1^*2^+3^+$&29\\
29&7.76&$2^+3^+4^*$&\hl{\textbf{30}}&7.04&$1^-3^-4^-$&0&7.71&$1^-2^+4^-$&0&7.76&$1^*2^+3^+$& \hl{\textbf{30}}\\
30&7.75&$2^+3^+4^*$&\hl{\textbf{30}}&7.07&$1^-3^-4^-$&0&7.71&$1^-2^+4^-$&0&7.75&$1^*2^+3^+$&\hl{\textbf{30}}\\

\bottomrule
\end{tabular}
\label{tbl:unCon_lam}
\end{table*}

We first scrutinise the algorithms' performance in the unconstrained case, where no constraints are imposed on the quality of the tours. We conduct the experiments on a complete graph with $n = 50$ nodes, and consider the following values for the other parameters: $\mu \in \{20, 21, \ldots, 30\}$. The termination criterion  and $\lambda$ are set to $5\times10^7$ diversity evaluations and $12$, respectively, based on the preliminary investigations. According to the pigeonhole principle, the most challenging case in unconstrained diversity optimisation occurs when $2n\mu$ is a factor of $|E|$ (the number of edges). While  For $n = 50$, the most difficult cases arise when $\mu$ is any factor of $24$. Thus, we set $\mu \in \{20, 21, \ldots, 30\}$ to take such challenging cases into account. 
\begin{table*}
%\footnotesize
\centering
\caption{Comparison of the proposed subset selection schemes and $(\mu+1)$~EA. The termination criterion is $10^7$ $H$-evaluations. The notations are in line with Table \ref{tbl:unCon_lam}}

\renewcommand{\tabcolsep}{7.5pt}
\renewcommand{\arraystretch}{0.9}
\begin{tabular}{lcr|cc|cc|cc|cc}
\toprule
         &&&\multicolumn{2}{c|}{ \textbf{EA}}&\multicolumn{2}{c|}{\textbf{Tournament}}&\multicolumn{2}{c|}{\textbf{Greedy}}&\multicolumn{2}{c}{$(\mu+1)$~\textbf{EA}}\\
\cmidrule(l{2pt}r{2pt}){4-5}
\cmidrule(l{2pt}r{2pt}){6-7}
\cmidrule(l{2pt}r{2pt}){8-9}
\cmidrule(l{2pt}r{2pt}){10-11}

             Instance&$\alpha$&$\mu$& $H$ & Stat (1) & $H$ & Stat (2) & $H$ & Stat (3) & $H$ & Stat (4) \\

\midrule
st70&0.05&50&5.67&$2^-3^+4^+$&\hl{\textbf{5.7}}&$1^+3^+4^+$&5.64&$1^-2^-4^+$&5.62&$1^-2^-3^-$\\
st70&0.05&100&5.7&$2^*3^+4^+$&\hl{\textbf{5.73}}&$1^*3^+4^+$&5.65&$1^-2^-4^*$&5.65&$1^-2^-3^*$\\
st70&0.12&50&\hl{\textbf{5.98}}&$2^+3^*4^+$&5.95&$1^-3^-4^+$&5.97&$1^*2^+4^+$&5.93&$1^-2^-3^-$\\
st70&0.12&100&\hl{\textbf{6.01}}&$2^+3^+4^+$&5.98&$1^-3^*4^+$&5.98&$1^-2^*4^+$&5.96&$1^-2^-3^-$\\
st70&0.25&50&\hl{\textbf{6.28}}&$2^+3^*4^+$&6.2&$1^-3^-4^-$&\hl{\textbf{6.28}}&$1^*2^+4^+$&6.25&$1^-2^+3^-$\\
st70&0.25&100&\hl{\textbf{6.3}}&$2^+3^*4^+$&6.25&$1^-3^-4^-$&\hl{\textbf{6.3}}&$1^*2^+4^+$&6.28&$1^-2^+3^-$\\
eil101&0.05&50&6.02&$2^-3^+4^+$&\hl{\textbf{6.04}}&$1^+3^+4^+$&5.97&$1^-2^-4^+$&5.93&$1^-2^-3^-$\\
eil101&0.05&100&6.05&$2^-3^+4^+$&\hl{\textbf{6.06}}&$1^+3^+4^+$&5.98&$1^-2^-4^*$&5.97&$1^-2^-3^*$\\
eil101&0.12&50&\hl{\textbf{6.35}}&$2^+3^+4^+$&6.31&$1^-3^-4^*$&6.34&$1^-2^+4^+$&6.3&$1^-2^*3^-$\\
eil101&0.12&100&\hl{\textbf{6.37}}&$2^+3^+4^+$&6.34&$1^-3^*4^+$&6.34&$1^-2^*4^+$&6.33&$1^-2^-3^-$\\
eil101&0.25&50&\hl{\textbf{6.73}}&$2^+3^*4^+$&6.63&$1^-3^-4^-$&\hl{\textbf{6.73}}&$1^*2^+4^+$&6.7&$1^-2^+3^-$\\
eil101&0.25&100&\hl{\textbf{6.75}}&$2^+3^+4^+$&6.68&$1^-3^-4^-$&6.74&$1^-2^+4^+$&6.73&$1^-2^+3^-$\\
a280&0.05&50&6.87&$2^-3^+4^+$&\hl{\textbf{6.88}}&$1^+3^+4^+$&6.84&$1^-2^-4^+$&6.82&$1^-2^-3^-$\\
a280&0.05&100&6.89&$2^-3^+4^+$&\hl{\textbf{6.9}}&$1^+3^+4^+$&6.85&$1^-2^-4^*$&6.84&$1^-2^-3^*$\\
a280&0.12&50&\hl{\textbf{7.15}}&$2^+3^+4^+$&7.11&$1^-3^-4^+$&7.13&$1^-2^+4^+$&7.1&$1^-2^-3^-$\\
a280&0.12&100&\hl{\textbf{7.16}}&$2^+3^+4^+$&7.15&$1^-3^*4^+$&7.14&$1^-2^*4^+$&7.14&$1^-2^-3^-$\\
a280&0.25&50&\hl{\textbf{7.49}}&$2^+3^+4^+$&7.4&$1^-3^-4^-$&7.48&$1^-2^+4^+$&7.45&$1^-2^+3^-$\\
a280&0.25&100&7.45&$2^*3^-4^-$&7.45&$1^*3^-4^-$&\hl{\textbf{7.49}}&$1^+2^+4^*$&7.48&$1^+2^+3^*$\\
\bottomrule
\end{tabular}
\label{tbl:Con_lam}
\end{table*}
% \aneta{In the Table 4 are two 25 entry. The description is not clear. It would be great to improve it. }
Table \ref{tbl:unCon_lam} summarises the results for these experiments. One can observe that the EA-based subset selection algorithm outperforms the competitors followed by the conventional $(\mu+1)$~EA and the greedy algorithm. The EA algorithm always hits $H_{\max}$ on all instances except where $\mu \in \{24, 25\}$. In these cases, the EA selection hits the optimum, 10 and 16 times, respectively. The $(\mu+1)$~EA cannot bring about the optimal population when $\mu\in\{23, 24, 25\}$ out of 30 runs and it hits the optimum only one time when $\mu = 26$. The $(\mu+1)$~EA performance gets better as we deviate from $\mu = 24$, and it results in the optimum for the two largest $\mu$ values in all 30 independent runs, similar to the EA algorithm. Moreover, the statistical observation confirms a significant difference between the EA algorithm and the $(\mu+1)$~EA in cases $\mu \in \{23, \ldots, 25\}$. The tournament and the greedy algorithms never resulted in the optimums values in this setting. In fact, the tournament algorithm's results are far away from the optimums. 

In conclusion, we can confirm that an offspring pool of $\lambda \geq 2$ can boost the algorithm performance in the unconstrained optimisation if a suitable subset selection method such as the EA-based approach is employed. %Next, we analyse the methods where a quality constraint is imposed to the problem.       

\subsection{Constrained Diversity Optimisation}
We now conduct a series experiments to evaluate the proposed methods' performance in constrained diversity optimisation. In the first set of experiments which the results are summarised in Table~\ref{tbl:Con_lam}, we test the algorithms on the instances st70, eil101, and a 280 from the TSPlib~\cite{Reinelt91tsplib}. We set the tournament size to $r=3$, the offspring pool size to $\lambda = 50$, and experiments with several values for the quality constraint $\alpha \in \{0.05, 0.12, 0,25\}$ 
% \jakob{In Figure~\ref{fig:boxplot} we also see results for $\alpha = 1$ and $\alpha = 3$. However, no results for these values are reported in the tables.}
and the population size $\mu \in \{50, 100\}$. The termination criterion is set to $10^7$ $H$-evaluations.
\begin{figure}
\centering
\includegraphics[width=\columnwidth]{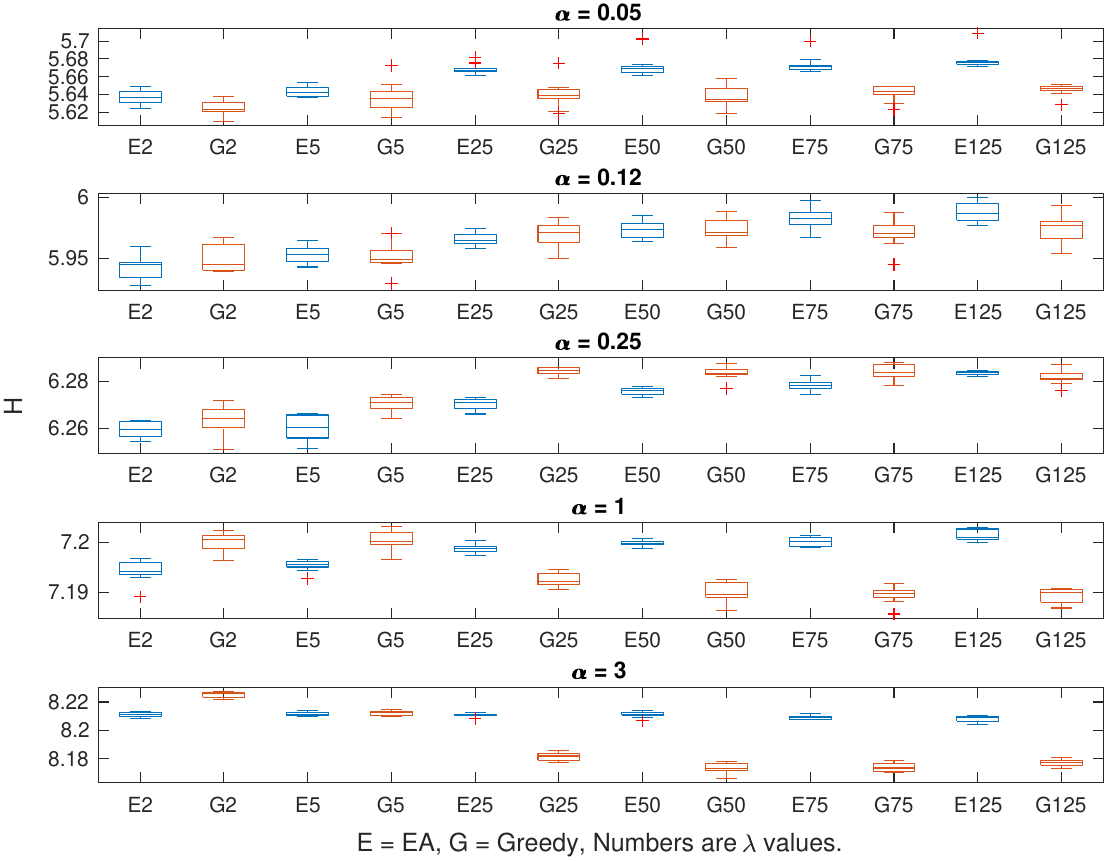}
% \squeezeup
% \squeezeup
% \squeezeup
\caption{Distributions of $H$-values of the final populations based on 10 independent runs for $(\mu+\lambda)$~EA (E) and Greedy (G) subset selection on instance st70. The plots show results for $\alpha \in \{0.02, 0.12, 0.25, 1, 3\}$ (row-wise) and $\lambda \in \{2, 5, 50, 125\}$ (from left to right).}
\label{fig:boxplot}
% \squeezeup
% \squeezeup
% \squeezeup
\end{figure} Table~\ref{tbl:Con_lam} presented in this section reveals some interesting findings. Firstly, the tournament selection method outperforms the other methods when the smallest considered $\alpha$-value ($\alpha = 0.05$) is used, with the EA algorithm following closely behind. Conversely, the vanilla $(\mu+1)$~EA performs the worst in these cases. This may be due to its inferior exploration capabilities, as k-tournament selection encourages exploration over exploitation. As $\alpha$ increases to 0.12, the EA-based selection method produces the best results, followed by the greedy and tournament methods. The greedy algorithm produces sets with slightly higher diversity when compared to tournament selection. However, the $(\mu+1)$~EA performs poorly in this case. For $\alpha = 0.25$, the greedy algorithm performs best for $\mu = 50$, while the EA outperforms the others for $\mu = 100$, with the $(\mu+1)$~EA ranking closely behind. The tournament selection method performs worst in these cases. Overall, the EA exhibits the most stable performance across all instances, always ranking as the best or second-best method. This may be attributed to the balance between exploration and exploitation that the EA offers. Selection methods with more exploration appear to perform better with tight quality constraints, whereas algorithms that require more exploitation perform better when $\alpha$ is large.

\begin{figure}
\centering
\includegraphics[width=\columnwidth]{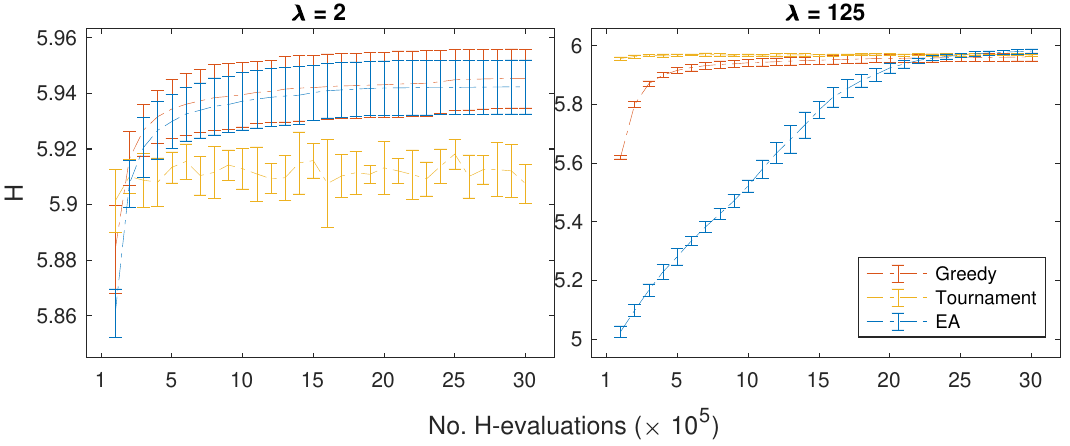}
% \squeezeup
% \squeezeup
% \squeezeup
\caption{Representative trajectories of the proposed subset selection methods on instance st70 with $\alpha = 0.12$ and $\lambda \in \{2, 125\}$.}
\label{fig:trend}
% \squeezeup
% \squeezeup
% \squeezeup
\end{figure}
In the next series of experiments, we investigate the performance of the EA and greedy methods on st70 when $\alpha \in \{0.02, 0.12, 0.25, 1, 3\}$ and $\lambda \in \{2, 5, 50, 125\}$. Figure~\ref{fig:boxplot} shows that the performance of the EA is slightly better for larger $\lambda$-values ($75, 125$), while it strongly depends on the $\alpha$-value for the greedy approach. The greedy methods preforms slightly better with large $\lambda$ when $\alpha$ is small ($0.05$), it is the other way around in the cases of large $\alpha$-values ($\alpha = 3$). As $\lambda \rightarrow 1$, the greedy algorithm becomes equivalent to the vanilla $(\mu+1)$~EA. We observed that $(\mu+1)$~EA outperforms the greedy method in unconstrained diversity. One could guess that smaller values of $\lambda$ works better for the greedy method for relaxed constraints. In general, the EA's performance is more stable and gets less affected by different values of $\alpha$, $\mu$, and $\lambda$.   

Figure \ref{fig:trend} illustrates representative trajectories of the proposed selection methods on st70 where $\alpha = 0.12$ and $\lambda = \{2, 125\}$. The tournament selection has the fastest convergence, and increase in $\lambda$ affects the convergence pace less compared to the other methods. The increase in $\lambda$ has the most significant impact on  the EA's convergence pace since there is linear relation between $L$ and $\lambda$ (we set $L = 2\mu\lambda$). Therefore, when $\lambda$ increases, more $H$-evaluations are spent on each loop of the EA selection. The increase in $\lambda$ also makes an impact on the greedy convergence pace, but not as severely as it does on the EA selection. 

\section*{Conclusion} \label{sec:con}

We studied survival selection methods for EDO working with an offspring population size greater than $1$. 
%We investigated EDO where the pool of offspring individuals is greater than~1. 
%Due to the dependency of individuals in the diversity calculation, most studies used the standard $(\mu+1)$~EA where at most one individual is replaced in each generation. 
We introduced three subset selection methods that enable us replacing more than one individuals with offspring for the next generation.
%, greedy subset selection, tournament subset selection, and EA-based subset selection. 
Our results show that an EA-based method as a sub-procedure of a $(\mu+\lambda)$~EA with $\lambda \geq 2$ outperforms the standard $(\mu+1)$~EA and the other proposed methods in most cases.
%Future work will focus on theoretical investigations of the proposed methods.

\section*{Acknowledgements}
This work was supported by the Australian Research Council through grant FT200100536.

%\frank{can remove some references}
\small
\bibliographystyle{IEEEtran}
\bibliography{bib}

@String{Computing = "Computing" }

@String{Computer = "{IEEE} Computer" }

@String{Springer = "Springer-Verlag" }

@Article{Reinelt91tsplib,
  author      = {Gerhard Reinelt},
  journal     = {ORSA Journal on Computing},
  title       = {{TSPLIB}--A Traveling Salesman Problem Library},
  year        = {1991},
  number      = {4},
  pages       = {376--384},
  volume      = {3}
}

@inproceedings{neumannGECCO23,
  title={Evolutionary Diversity Optimization for the Detection and Concealment of Spatially Defined Communication Networks},
  author={Aneta Neumann and Sharlotte Gounder and Xiankun Yan and Gregory Sherman and  Benjamin Campbell and Mingyu Guo and Frank Neumann},
  booktitle={{GECCO}},
  year={2023},
  publisher = {{ACM}}
}

@inproceedings{alexander2017evolution,
  author       = {Bradley Alexander and
                  James Kortman and
                  Aneta Neumann},
  title        = {Evolution of artistic image variants through feature based diversity
                  optimisation},
  booktitle    = {{GECCO}},
  pages        = {171--178},
  publisher    = {{ACM}},
  year         = {2017}
}

@inproceedings{ulrich2011maximizing,
  author       = {Tamara Ulrich and
                  Lothar Thiele},
  title        = {Maximizing population diversity in single-objective optimization},
  booktitle    = {{GECCO}},
  pages        = {641--648},
  publisher    = {{ACM}},
  year         = {2011}
}

@inproceedings{neumann2018discrepancy,
  author       = {Aneta Neumann and
                  Wanru Gao and
                  Carola Doerr and
                  Frank Neumann and
                  Markus Wagner},
  title        = {Discrepancy-based evolutionary diversity optimization},
  booktitle    = {{GECCO}},
  pages        = {991--998},
  publisher    = {{ACM}},
  year         = {2018}
}

@inproceedings{neumann2019evolutionary,
  author       = {Aneta Neumann and
                  Wanru Gao and
                  Markus Wagner and
                  Frank Neumann},
  title        = {Evolutionary diversity optimization using multi-objective indicators},
  booktitle    = {{GECCO}},
  pages        = {837--845},
  publisher    = {{ACM}},
  year         = {2019}
}

@inproceedings{bossek2019evolving,
  author       = {Jakob Bossek and
                  Pascal Kerschke and
                  Aneta Neumann and
                  Markus Wagner and
                  Frank Neumann and
                  Heike Trautmann},
  title        = {Evolving diverse {TSP} instances by means of novel and creative mutation
                  operators},
  booktitle    = {{FOGA}},
  pages        = {58--71},
  publisher    = {{ACM}},
  year         = {2019}
}

@article{doi:10.1162/evcoa00274,
author       = {Wanru Gao and
                  Samadhi Nallaperuma and
                  Frank Neumann},
  title        = {Feature-Based Diversity Optimization for Problem Instance Classification},
  journal      = {Evol. Comput.},
  volume       = {29},
  number       = {1},
  pages        = {107--128},
  year         = {2021}
}

@inproceedings{viet2020evolving,
  author       = {Anh Viet Do and
                  Jakob Bossek and
                  Aneta Neumann and
                  Frank Neumann},
  title        = {Evolving diverse sets of tours for the travelling salesperson problem},
  booktitle    = {{GECCO}},
  pages        = {681--689},
  publisher    = {{ACM}},
  year         = {2020}
}

@article{nagata2020high,
  author       = {Yuichi Nagata},
  title        = {High-Order Entropy-Based Population Diversity Measures in the Traveling
                  Salesman Problem},
  journal      = {Evol. Comput.},
  volume       = {28},
  number       = {4},
  pages        = {595--619},
  year         = {2020}
}

@inproceedings{NikfarjamBN021a,
  author    = {Adel Nikfarjam and
               Jakob Bossek and
               Aneta Neumann and
               Frank Neumann},
  title     = {Entropy-based evolutionary diversity optimisation for the traveling
               salesperson problem},
  booktitle = {{GECCO}},
  pages     = {600--608},
  publisher = {{ACM}},
  year      = {2021}
}

@inproceedings{NikfarjamB0N21b,
  author    = {Adel Nikfarjam and
               Jakob Bossek and
               Aneta Neumann and
               Frank Neumann},
  title     = {Computing diverse sets of high quality {TSP} tours by EAX-based evolutionary
               diversity optimisation},
  booktitle = {{FOGA}},
  pages     = {9:1--9:11},
  publisher = {{ACM}},
  year      = {2021}
}

@inproceedings{Bossek021tree,
  author    = {Jakob Bossek and
               Frank Neumann},
  title     = {Evolutionary diversity optimization and the minimum spanning tree
               problem},
  booktitle = {{GECCO}},
  pages     = {198--206},
  publisher = {{ACM}},
  year      = {2021}
}

@inproceedings{NeumannB021,
  author    = {Aneta Neumann and
               Jakob Bossek and
               Frank Neumann},
  title     = {Diversifying greedy sampling and evolutionary diversity optimisation
               for constrained monotone submodular functions},
  booktitle = {{GECCO}},
  pages     = {261--269},
  publisher = {{ACM}},
  year      = {2021}
}

@inproceedings{NikTTPEDO,
  author    = {Adel Nikfarjam and
               Aneta Neumann and
               Frank Neumann},
  title     = {Evolutionary diversity optimisation for the traveling thief problem},
  booktitle = {{GECCO}},
  pages     = {749--756},
  publisher = {{ACM}},
  year      = {2022}
}

@inproceedings{CullyM13,
  author    = {Antoine Cully and
               Jean{-}Baptiste Mouret},
  title     = {Behavioral repertoire learning in robotics},
  booktitle = {{GECCO}},
  pages     = {175--182},
  publisher = {{ACM}},
  year      = {2013}
}

@inproceedings{CluneML13,
  author    = {Jeff Clune and
               Jean{-}Baptiste Mouret and
               Hod Lipson},
  title     = {Summary of "the evolutionary origins of modularity"},
  booktitle = {{GECCO} (Companion)},
  pages     = {23--24},
  publisher = {{ACM}},
  year      = {2013}
}

@inproceedings{PughSSS15,
  author    = {Justin K. Pugh and
               Lisa B. Soros and
               Paul A. Szerlip and
               Kenneth O. Stanley},
  title     = {Confronting the Challenge of Quality Diversity},
  booktitle = {{GECCO}},
  pages     = {967--974},
  publisher = {{ACM}},
  year      = {2015}
}

@article{PughSS16,
  author    = {Justin K. Pugh and
               Lisa B. Soros and
               Kenneth O. Stanley},
  title     = {Quality Diversity: {A} New Frontier for Evolutionary Computation},
  journal   = {Frontiers Robotics {AI}},
  volume    = {3},
  pages     = {40},
  year      = {2016}
}

@inproceedings{NikfarjamDN22,
  author    = {Adel Nikfarjam and
               Anh Viet Do and
               Frank Neumann},
  title     = {Analysis of Quality Diversity Algorithms for the Knapsack Problem},
  booktitle = {{PPSN} {(2)}},
  series    = {Lecture Notes in Computer Science},
  volume    = {13399},
  pages     = {413--427},
  publisher = {Springer},
  year      = {2022}
}

@inproceedings{Bossek022,
  author    = {Jakob Bossek and
               Frank Neumann},
  title     = {Exploring the feature space of {TSP} instances using quality diversity},
  booktitle = {{GECCO}},
  pages     = {186--194},
  publisher = {{ACM}},
  year      = {2022}
}

@inproceedings{TjanakaFTN22,
  author    = {Bryon Tjanaka and
               Matthew C. Fontaine and
               Julian Togelius and
               Stefanos Nikolaidis},
  title     = {Approximating gradients for differentiable quality diversity in reinforcement
               learning},
  booktitle = {{GECCO}},
  pages     = {1102--1111},
  publisher = {{ACM}},
  year      = {2022}
}

@inproceedings{EarleSFNT22,
  author    = {Sam Earle and
               Justin Snider and
               Matthew C. Fontaine and
               Stefanos Nikolaidis and
               Julian Togelius},
  title     = {Illuminating diverse neural cellular automata for level generation},
  booktitle = {{GECCO}},
  pages     = {68--76},
  publisher = {{ACM}},
  year      = {2022}
}

@article{DoGNN22,
  author       = {Anh Viet Do and
                  Mingyu Guo and
                  Aneta Neumann and
                  Frank Neumann},
  title        = {Analysis of Evolutionary Diversity Optimization for Permutation Problems},
  journal      = {{ACM} Trans. Evol. Learn. Optim.},
  volume       = {2},
  number       = {3},
  pages        = {11:1--11:27},
  year         = {2022}
}

@inproceedings{DoerrLMN17,
  author       = {Benjamin Doerr and
                  Huu Phuoc Le and
                  R{\'{e}}gis Makhmara and
                  Ta Duy Nguyen},
  title        = {Fast genetic algorithms},
  booktitle    = {{GECCO}},
  pages        = {777--784},
  publisher    = {{ACM}},
  year         = {2017}
}

@inproceedings{NIKSAT,
  title={Evolutionary Diversity Optimisation in Constructing  Satisfying Assignments},
  author={Adel Nikfarjam and Ralf Rothenberger and Frank Neumann and Tobias Friedrich},
  booktitle={{GECCO}},
  year={2023},
  publisher = {{ACM}}
}

@article{baraton2025bayesian, title={Bayesian Quality-Diversity optimization forconditional search-space problems}, author={Baraton, Lucas and Urbano, Annafedericaand Brevault, Loic and Balesdent, Mathieu}, journal={Optimization and Engineering},pages={1--46}, year={2025}, publisher={Springer} }

@inproceedings{NikfarjamSND024,
  author       = {Adel Nikfarjam and
                  Ty Stanford and
                  Aneta Neumann and
                  Dorothea Dumuid and
                  Frank Neumann},
  title        = {Quality Diversity Approaches for Time-Use Optimisation to Improve
                  Health Outcomes},
  booktitle    = {{GECCO}},
  publisher    = {{ACM}},
  year         = {2024}
}

@article{LiEDE17,
  author    = {Xiaodong Li and
               Michael G. Epitropakis and
               Kalyanmoy Deb and
               Andries P. Engelbrecht},
  title     = {Seeking Multiple Solutions: An Updated Survey on Niching Methods and
               Their Applications},
  journal   = {{IEEE} Trans. Evol. Comput.},
  volume    = {21},
  number    = {4},
  pages     = {518--538},
  year      = {2017}
}

@article{GuBLQ24,
  author       = {Yu{-}Ran Gu and
                  Chao Bian and
                  Miqing Li and
                  Chao Qian},
  title        = {Subset Selection for Evolutionary Multiobjective Optimization},
  journal      = {{IEEE} Trans. Evol. Comput.},
  volume       = {28},
  number       = {2},
  pages        = {403--417},
  year         = {2024}
}

\end{document}